%% file: main_csa_sa.tex
\documentclass[10pt, a4paper, notitlepage]{article}
\usepackage[utf8]{inputenc}
\usepackage{physics}
\usepackage{amsmath,amssymb}
\usepackage{mathtools,bbm}
\usepackage{graphicx}
\usepackage{hyperref,cleveref}
\usepackage[title]{appendix}
\usepackage{aligned-overset}
\usepackage{bm}
\usepackage{algorithmicx,algorithm,algpseudocode}
\usepackage{placeins}
\usepackage{tikz}
\usepackage{subcaption}

\usepackage{geometry}
\input{cmds.tex}

\hypersetup{colorlinks=true, citecolor=black, linkcolor=black}
\crefformat{equation}{(#2#1#3)}
\crefformat{section}{#2#1#3}
\crefformat{figure}{#2#1#3}
\crefformat{algorithm}{#2#1#3}
\crefformat{table}{#2#1#3}

\newcommand\mywidth{7cm}

\begin{document}
	
	\title{\large Technical Report \\[2ex] 
	\LARGE Mutation Strength Adaptation of the $\muilam{\mu}{\lambda}$-ES for Large Population Sizes on the Sphere Function}
	\author{Amir Omeradzic and Hans-Georg Beyer}
	\date{\today}
	\maketitle

	\begin{abstract}
		The mutation strength adaptation properties of a multi-recombinative $\muilam{\mu}{\lambda}$-ES are studied for isotropic mutations.
		To this end, standard implementations of cumulative step-size adaptation (CSA) and mutative self-adaptation ($\sigma$SA) are investigated experimentally and theoretically by assuming large population sizes ($\mu$) in relation to the search space dimensionality ($N$).
		The adaptation is characterized in terms of the scale-invariant mutation strength on the sphere in relation to its maximum achievable value for positive progress.
			Standard CSA-variants show notably different adaptation properties and progress rates on the sphere, becoming slower or faster as $\mu$ or $N$ are varied.
			This is shown by investigating common choices for the cumulation and damping parameters. 
			Standard $\sigma$SA-variants (with default learning parameter settings) can achieve faster adaptation and larger progress rates compared to the CSA.
			However, it is shown how self-adaptation affects the progress rate levels negatively.
			Furthermore, differences regarding the adaptation and stability of $\sigma$SA with log-normal and normal mutation sampling are elaborated.
	\end{abstract}
	
	\input{intro.tex}

	\input{content.tex}
	\input{conclusion.tex}

	\subsubsection*{Acknowledgments}
	This work was supported by the Austrian Science Fund (FWF) under grant P33702-N.

	

\input{main_csa_sa.bbl}
	\bibliographystyle{alpha}
	
\end{document}

%% file: cmds.tex
\newcommand{\mycomment}[1]{}

\newcommand{\ParR}[1]{\left(#1\right)}

\newcommand{\muilam}[2]{(#1/#1_I,#2)}

\newcommand{\SUMM}[0]{\sum_{m=1}^{\mu}}

\newcommand{\CMULAM}[0]{c_{\mu/\mu,\lambda}}
\newcommand{\CTHETA}[0]{c_{\vartheta}}
\newcommand{\CTHETAlong}[0]{\ONETWOPI \frac{1}{\vartheta} \EXP{ -\frac{1}{2} \ParR{\Phi^{-1}(\vartheta)}^2 }}

\newcommand{\ONETWOPI}[1][]{\frac{1}{\sqrt{2\pi}#1}}

\newcommand{\e}[0]{\mathrm{e}}
\newcommand{\EXP}[1]{\operatorname{exp}\left[ #1 \right]}

\newcommand{\EV}[1]{\operatorname{E}\left[#1\right]}

\newcommand{\vt}[0]{\vartheta}

\newcommand{\signss}[0]{\sigma^*_{\mathrm{ss}}}

\newcommand{\signzero}[0]{\sigma^*_{\varphi_0}}

\newcommand{\rec}[1]{\langle#1\rangle}

\newcommand{\SALN}[0]{\sigma\text{SA}_L}
\newcommand{\SAEP}[0]{\sigma\text{SA}_N}
\newcommand{\cs}[0]{c_\sigma}
\newcommand{\gvec}[2]{\vb{#1}^{(#2)}}
\newcommand{\bcoeff}[0]{\frac{1}{\cs D/(1-\cs) + \sqrt{2}\CTHETA D/\sqrt{N}}}
\newcommand{\snorm}[1]{||\vb{s}^{(#1)}||}
\newcommand{\znorm}[1]{||\rec{\vb{z}}^{(#1)}||}


%% file: intro.tex
\section{Introduction}
The population size of a multi-recombinative Evolution Strategy (ES) is crucial to improve the search behavior on noisy test functions \cite{Bey00b,AB00d} and highly multimodal functions with adequate global structure \cite{HK04,SB24,OB24}.
Besides the population size, the adaptation properties of the mutation strength $\sigma$-adaption also significantly influence the search behavior of the ES.
State-of-the-art $\sigma$-adaptation methods are cumulative step-size adaptation (CSA, see \cite{Han98,HO01,Arn02,hansen2023cma}) and mutative self-adaptation ($\sigma$SA, see \cite{Sch77,Bey00b,Mey07,OB24a}).
The goal of this paper is to gain deeper insight into the adaptation properties of CSA and $\sigma$SA as a function of the population size $\mu$ and search space dimensionality $N$.
To this end, first experimental and theoretical analyses will be conducted on the sphere function.
While this is a simplification of real-world, more complex optimization problems, it will allow to gain better understanding of the basic adaptation properties as $\mu$ or $N$ are varied.

The presented analysis is mainly motivated as a building block for a future goal to study adaptive (online) population size control on ES, where the population size is changed dynamically depending on the current ES-performance to improve its search under noise and multimodality.
First tests on standard CSA- and $\sigma$SA-implementations have revealed significant differences of the adaptation speed on simple test functions such as the sphere.
These differences have a notable impact on the performance of regular ES working at constant population size.
Hence, in order to understand ES with dynamic population control, the first step is an analysis of the underlying $\sigma$-adaptation behavior.

\input{alg_es_csa.tex}
\input{algo_sigSA_v2.tex}
%
Algorithm~\cref{alg:csa} shows the implementation of a $\muilam{\mu}{\lambda}$-CSA-ES.
It works by generating a cumulation path \cref{eq:csa_s} of selected (recombined) mutation directions.
Then, the update rules \cref{eq:new_han_v1} or \cref{eq:neq_han_v2}, respectively, are applied to control $\sigma$.
Different CSA-parametrizations will be discussed in Sec.~\cref{sec:gamma_csa}.
Algorithm~\cref{alg:sa} shows a $\muilam{\mu}{\lambda}$-$\sigma$SA-ES with mutative self-adaptation.
A standard self-adaptive ES samples log-normally distributed mutation strengths (denoted by $\SALN$) for each offspring \cite{Sch77, Bey00b}.
The sampling is controlled using the learning parameter $\tau$. 
Offspring mutation strengths are generated according to
\begin{align}\begin{split}\label{eq:metaep_logn}
		\widetilde{\sigma}_L = \sigma \e^{\tau\mathcal{N}(0,1)}.
\end{split}\end{align}
Selection by fitness implicitly selects suitable mutation strengths, which are recombined to obtain a new parental $\sigma$.
As an alternative of sampling log-normally distributed values, one can introduce a normal sampling scheme (denoted by $\SAEP$) as \cite{Bey00b,OB24a}
\begin{align}\begin{split}\label{eq:metaep_ep1}
		\widetilde{\sigma}_N = \sigma \qty(1+\tau\mathcal{N}(0,1)).
\end{split}\end{align}
The expected values of the sampling schemes \cref{eq:metaep_logn} and \cref{eq:metaep_ep1} can be evaluated as
\begin{align}
	\EV{\widetilde{\sigma}_L} &= \sigma\e^{\tau^2/2} \label{eq:metaep_logn2} \\
	\EV{\widetilde{\sigma}_N} &= \sigma. \label{eq:metaep_ep2}
\end{align}
The former yields a biased sampling of mutation strengths since the expected value (under random selection) is larger than the initial value.
The latter sampling is referred to as unbiased.
Details on the bias property can be found in \cite{OB24a}.
It will explain some of the later observed differences between $\SALN$ and $\SAEP$.

In Sec.~\cref{sec:phi}, the sphere progress rate for large populations is derived.
Furthermore, first experiments comparing the progress rates of CSA and $\sigma$SA are conducted.
In Sec.~\cref{sec:gamma_csa}, a more detailed analysis of the CSA-ES on the sphere will be presented.
To this end, theoretical and experimental investigations will be conducted.
Thereafter, in Sec.~\cref{sec:gamma_sa}, self-adaptive $\sigma$SA-ES using log-normal and normal mutation sampling are studied on the sphere.
Finally, conclusions are drawn in Sec.~\cref{sec:conc}.

\section{Sphere Progress Rate for Large Populations}
\label{sec:phi}
\begin{figure}[t]
	\centering
	\includegraphics[width=\mywidth]{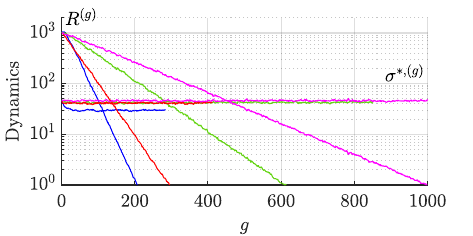}
	\includegraphics[width=\mywidth]{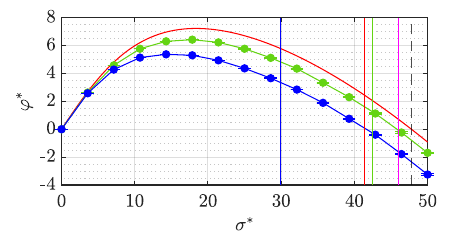}
	\caption{Median dynamics on the sphere (10 trials) for $\mu=100$, $\lambda=200$, and $N=100$.
		Given the $R$-dynamics, the progress rate is measured by evaluating $\varphi^{*,(g)} = (R^{(g)}-R^{(g+1)})\frac{N}{R^{(g)}}$, and averaged as $\varphi^*_\mathrm{meas}= \mathrm{mean}(\varphi^{*,(g_0:g_\mathrm{end})})$ ($g_0 \leq g \leq g_\mathrm{end}$, $g_0=20$ reducing initialization effects).
		On the left, one has CSA~\cref{eq:sqrtN} (red, $\varphi^*_\mathrm{meas}\approx2.3$), CSA~\cref{eq:linN} (magenta, $\varphi^*_\mathrm{meas}\approx0.7$), $\SALN$ with $\tau=1/\sqrt{2N}$ (blue, $\varphi^*_\mathrm{meas}\approx3.5$), and $\SALN$  with $\tau=1/\sqrt{8N}$ (green, $\varphi^*_\mathrm{meas}\approx1.1$).
		On the right, one has $\varphi^*$ from \cref{eq:sph_Ndep_pc} (solid red). 
		For the self-adaptive ES, $\varphi^*(\sigma^*,\tau)=(R^{(0)}-R^{(1)})\frac{N}{R^{(0)}}$ was determined using one-generation experiments with $10^4$ trials for $\tau=1/\sqrt{8N}$ (green) and $\tau=1/\sqrt{2N}$ (blue).
		The vertical lines mark measured steady-state $\signss$ (same color code as on the right). 
		The second zero of \cref{eq:sph_Ndep_pc} is marked in dashed black.}
	\label{fig:dyn_phi_csa_sa}
\end{figure}

The sphere function is defined as $f(R) \coloneqq R^2$, $R = ||\vb{y}||$, $\vb{y}\in\mathbb{R}^N$.
The progress rate is defined as the expected change of the residual distance between two generations $g$ as
\begin{align}\begin{split}\label{sec:dyn_phidef}
		\varphi \coloneqq R^{(g)}-\EV{R^{(g+1)}}.
	\end{split}
\end{align}
Due to the scale-invariance on the sphere, one can define a normalized progress rate $\varphi^*$ and a scale-invariant (normalized) mutation strength $\sigma^*$ according to
\begin{align}\begin{split}\label{eq:sign} 
		\varphi^* = \varphi N / R,\quad \sigma^* = \sigma N/R.
\end{split}\end{align}
After the initialization phase has passed, the CSA-ES and $\sigma$SA-ES realize a constant $\sigma^*$-level and constant positive progress $\varphi^*>0$ (in expectation) on the sphere, see also Fig.~\cref{fig:dyn_phi_csa_sa}.
The sphere progress rate is a known quantity.
The (normalized) progress rate of the sphere derived in \cite[(6.54)]{Bey00b} with progress coefficient $\CMULAM$ is given by
\begin{align}\begin{split}\label{eq:sph_Ndep_pc}
		\varphi^* &= \frac{\CMULAM\sigma^*(1 + \sigma^{*2}/2\mu N)}{\sqrt{1+\sigma^{*2}/\mu N}\sqrt{1+\sigma^{*2}/2N}}
		-N\qty(\sqrt{1+\sigma^{*2}/\mu N}-1) + O\qty(N^{-1/2}).
\end{split}\end{align}
The numerically obtained (non-trivial) zero of \cref{eq:sph_Ndep_pc} will be denoted as $\sigma^*_0$
\footnote{For $N\lessapprox100$ it is advisable to calculate $\varphi^*$ from one-generation experiments of \cref{sec:dyn_phidef} using normalization \cref{eq:sign}.
Since formula \cref{eq:sph_Ndep_pc} neglects terms of $O\qty(N^{-1/2})$, higher accuracy of $\sigma^*_0$ is achieved with one-generation simulations.
This is necessary for experiments at very slow adaptation with $\signss\lessapprox\sigma^*_0$ (at low $N$).
}.
The analytic second zero (to be derived) will be denoted by $\signzero$.
An important characteristic of the sphere progress rate is that there is a range $\sigma^*\in(0,\sigma^*_0)$ (given $N$, parent population size $\mu$, and offspring population size $\lambda$) where positive progress $\varphi^*>0$ can be achieved.
Depending on how the $\sigma$-adaptation method is parameterized, the ES reaches a different steady-state $\sigma^*\in(0,\sigma^*_0)$.

Now, approximations are applied to \cref{eq:sph_Ndep_pc} assuming large population sizes (and constant truncation ratio $\vt=\mu/\lambda$).
Assuming $\mu N \gg \sigma^{*2}$, one can apply the Taylor-expansion $\sqrt{1+\sigma^{*2}/\mu N} = 1 + \sigma^{*2}/2\mu N + O( (\sigma^{*2}/\mu N)^2)$.
Furthermore, $\CMULAM\simeq\CTHETA$ is only a function of the truncation ratio $\vartheta$ \cite[(6.113)]{Bey00b}.
By neglecting higher order terms, one gets
\begin{align}\begin{split}\label{eq:sph_med}
		\varphi^* &\simeq \frac{\CTHETA\sigma^*}{\sqrt{1+\sigma^{*2}/2N}} - \frac{\sigma^{*2}}{2\mu}.
\end{split}\end{align}
Now we simplify \cref{eq:sph_med} by assuming $\sigma^*/2N\gg1$, such that ``1" is neglected within the square-root, which yields
\begin{equation}\label{sec:dyn_phi_large}
	\varphi^* \simeq \sqrt{2N}\CTHETA - \frac{\sigma^{*2}}{2\mu}.
\end{equation}
This is justified by assuming that comparably large mutation strengths $\sigma^{*}$ are realized, see also discussion of Fig.~\cref{fig:dyn_phi_csa_sa}.
The zero of approximation \cref{sec:dyn_phi_large} is given by
\begin{align}\begin{split}\label{eq:signzero_approx}
		\signzero = (8N)^{1/4}(\CTHETA\mu)^{1/2}.
\end{split}\end{align}
Now the approach is to characterize the $\sigma$-adaptation on the sphere in terms of a steady-state $\signss$ w.r.t.~the second zero $\signzero$, for which an analytic solution is available.
This is justified by the observation that for sufficiently slow adaptation the steady-state $\sigma^*=\signss$ lies in the vicinity of the second zero, such that $\signss\lessapprox\signzero$.
This will also be justified by experiments. 
Introducing a scaling factor $0<\gamma<1$ (slow-adaptation: $\gamma\lessapprox1$), one sets
\begin{align}\begin{split}\label{eq:signzero_approx_gam_v2}
		\signss = \gamma\signzero = \gamma(8N)^{1/4}(\CTHETA\mu)^{1/2}.
\end{split}\end{align}
Figure~\cref{fig:dyn_phi_csa_sa} shows example dynamics one the sphere (left) and the corresponding progress rate (right) to illustrate the approach \cref{eq:signzero_approx_gam_v2}.
First, note that different convergence rates (see generations $g$) are realized among the CSA- and $\sigma$SA-implementations (details of the implementations are given in Secs.~\cref{sec:gamma_csa} and \cref{sec:gamma_sa}).
Furthermore, they realize different steady-state $\signss$-levels.
Note that all algorithms (except $\sigma$SA with $\tau=1/\sqrt{2N}$) operate at relatively large $\signss$ close to $\sigma^*_0$.
The optimal value lies approximately at $\sigma^*\lessapprox20$.
To illustrate how the convergence rate is related to the progress rate and the $\sigma^*$-level, measured $\varphi^*_\mathrm{meas}$-values are given (see caption) and compared to $\varphi^*$-curves on the right.
To this end, one looks at the intersection of the curves with the corresponding vertical lines (same color scheme) and compares it with $\varphi^*_\mathrm{meas}$.
For the CSA (red and magenta lines), one observes very good agreement with \cref{eq:sph_Ndep_pc}.
For the $\sigma$SA, however, it is necessary to compare the measured progress rates to simulations of $\varphi^*(\sigma^*,\tau)$ with $\tau>0$.
Otherwise, comparably large deviations are observed.
Hence, similar $\signss$-levels of CSA~\cref{eq:sqrtN} (red) and $\sigma$SA~\cref{eq:metaep_logn} ($\tau=1/\sqrt{2N}$, green) yield to notably different progress and convergence rates observed on the left.

The experiments illustrate that \cref{eq:signzero_approx_gam_v2} is justified for sufficiently slow adaptation with $\gamma\lessapprox1$.
However, it also illustrates that $\tau>0$ introduces an error of the true (simulated) progress rate to the prediction \cref{eq:sph_Ndep_pc} which was derived assuming $\tau=0$.
Hence, \cref{eq:sph_Ndep_pc} serves as an approximation of the progress rate for the $\sigma$SA in the limit $\tau\rightarrow0$.
Furthermore, similar $\signss$-levels of CSA and $\sigma$SA yield different convergence rates.
Now that basic differences between the $\sigma$-adaptation methods have been studied, a more detailed analysis of the CSA and $\sigma$SA are given in the next two sections.

\begin{figure}[t]
	\centering
	\begin{subfigure}{\textwidth}
		\centering
		\includegraphics[width=\mywidth]{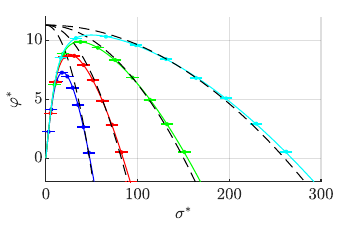}
		\includegraphics[width=\mywidth]{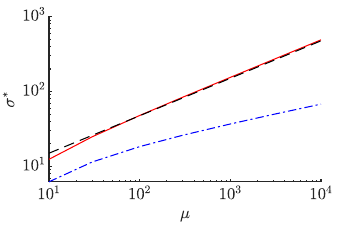}
		\caption{Variation of $\mu$ at $N=100$.
			On the left, exemplary $\varphi^*(\sigma^*)$ are shown for $\mu=100,300,1000,3000$.}
		\label{fig:phi}
	\end{subfigure}
	\begin{subfigure}{\textwidth}
		\centering
		\includegraphics[width=\mywidth]{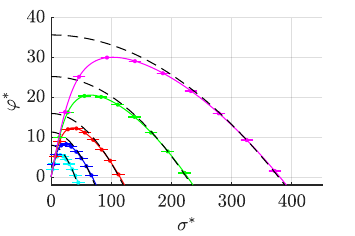}
		\includegraphics[width=\mywidth]{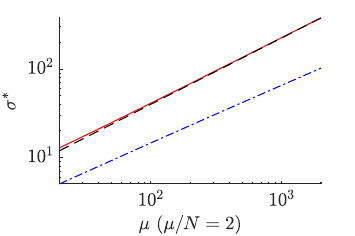}
		\caption{Variation of $\mu$ and $N$ with $\mu/N=2$.
			On the left, exemplary $\varphi^*(\sigma^*)$ are shown for $\mu=100,200,400,1000,2000$. }
		\label{fig:sign}
	\end{subfigure}
	\caption{Sphere progress rate $\varphi^*(\sigma^*)$ for varying population sizes.
		On the left, the solid lines show \cref{eq:sph_Ndep_pc} and the corresponding data points \cref{sec:dyn_phidef} averaged over $10^4$ trials and normalized using $\varphi^*=\varphi N/R$.
		The dashed line shows \cref{sec:dyn_phi_large}.
		On the right, \cref{eq:sph_Ndep_pc} is used to numerically calculate the second zero $\sigma^*_0$ (red solid) and $\hat{\sigma}^* = \operatorname{arg\,max}\varphi^*(\sigma^*)$ (dash-dotted blue).
		The black dashed line shows approximation \cref{eq:signzero_approx}.}
	\label{fig:phi_sign}
\end{figure}\noindent

%% file: alg_es_csa.tex
\begin{algorithm}[t]
	\caption{$(\mu/\mu_I, \lambda)$-CSA-ES}
	\label{alg:csa}
	\begin{algorithmic}[1]
		\State $g \gets 0$ 
		\State $\operatorname{initialize}\qty(\vb{y}^{(0)}, \vb{s}^{(0)}, \sigma^{(0)}, c_\sigma, d_\sigma, D)$ 
		\Repeat
		\For{$l = 1, ..., \lambda$} 
		\State $\tilde{\vb{z}}_l \gets [\mathcal{N}(0,1), ..., \mathcal{N}(0,1)]$
		\State $\tilde{\vb{y}}_l \gets \vb{y}^{(g)} + \sigma^{(g)}\tilde{\vb{z}}_l$ 
		\State $\tilde{f}_l \gets f(\vb{\tilde{y}}_l)$
		\EndFor
		\State	$(\tilde{f}_{1;\lambda}, ..., \tilde{f}_{m;\lambda}, ..., \tilde{f}_{\mu;\lambda} ) \gets \operatorname{sort}(\tilde{f}_1, ..., \tilde{f}_\lambda )$
		\State $\vb{y}^{(g+1)} \gets \frac{1}{\mu}\SUMM \tilde{\vb{y}}_{m;\lambda}$ 
		\State $\rec{\vb{z}}^{(g+1)} \gets \frac{1}{\mu}\SUMM \vb{z}_{m;\lambda}$
		\State $\vb{s}^{(g+1)} \gets (1-c_\sigma)\vb{s}^{(g)} + \sqrt{\mu c_\sigma(2-c_\sigma)}\rec{\vb{z}}^{(g+1)}$
		\State $\sigma^{(g+1)} \gets
		\begin{cases} 
			\sigma^{(g)}\EXP{\frac{1}{D}\qty(\norm{\vb{s}^{(g+1)}}/E_\chi-1)} \qq{via \cref{eq:new_han_v1}} \\
			\sigma^{(g)}\EXP{\frac{c_\sigma}{d_\sigma}\qty(\norm{\vb{s}^{(g+1)}}/E_\chi-1)} \qq{via \cref{eq:neq_han_v2}}
		\end{cases}$
		\State $g \gets g+1$
		\Until termination criterion
	\end{algorithmic}
\end{algorithm}%

%% file: algo_sigSA_v2.tex
\begin{algorithm}[t]
	\caption{$(\mu/\mu_I, \lambda)$-$\sigma$SA-ES with (log-)normal operator}
	\label{alg:sa}
	\begin{algorithmic}[1]
		\State \label{alg:line1}$g \gets 0$ 
		\State \label{alg:line2}$\operatorname{initialize}\big(\vb{y}^{(0)}, \sigma^{(0)}, \tau)$
		\Repeat \label{alg:line3}
		\For{$l = 1, ..., \lambda$} \label{alg:line4} 
		\State  \label{alg:line5} $\tilde{\sigma}_l \gets  
		\begin{cases} 
			\SALN: &\sigma^{(g)} \e^{\tau\mathcal{N}_l(0, 1)} \qq{via \cref{eq:metaep_logn}} \\  
			\SAEP: &\sigma^{(g)} \qty(1+\tau\mathcal{N}_l(0, 1))  \qq{via \cref{eq:metaep_ep1}}
		\end{cases}$
		\State  \label{alg:line6} $\tilde{\vb{y}}_l \gets \vb{y}^{(g)} + \tilde{\sigma}_l\mathcal{N}_l(\vb{0}, \vb{1})$ 
		\State  \label{alg:line7} $\tilde{f}_l \gets f(\vb{\tilde{y}}_l)$
		\EndFor \label{alg:line8}
		\State  \label{alg:line9} $(\tilde{f}_{1;\lambda}, ..., \tilde{f}_{m;\lambda}, ..., \tilde{f}_{\mu;\lambda} ) \gets \operatorname{sort}(\tilde{f}_1, ..., \tilde{f}_\lambda )$
		\State  \label{alg:line10} $\vb{y}^{(g+1)} \gets \frac{1}{\mu}\SUMM \tilde{\vb{y}}_{m;\lambda}$
		\State \label{alg:line11} $\sigma^{(g+1)} \gets \frac{1}{\mu}\SUMM \tilde{\sigma}_{m;\lambda}$
		\State \label{alg:line12} $g \gets g+1$
		\Until termination criterion
	\end{algorithmic}
\end{algorithm}

%% file: content.tex
\section{Cumulative Step-Size Adaptation}
\label{sec:gamma_csa}

The analysis of the CSA in this section consists of four parts.
In Sec.~\cref{sec:csa_intro}, the general algorithm for $\sigma$-adaptation via CSA is presented.
Furthermore, standard parameter sets for cumulation constant and damping are introduced and discussed.
Then, in Sec.~\cref{sec:csa_stst}, the derivation of the sphere steady-state evolution equations is presented.
The obtained equations need to be simplified using certain approximations in order to obtain closed-form solutions later.
This will be discussed throughout the section.
Afterwards in Sec.~\cref{sec:csa_iteration}, the steady-state of the CSA is analyzed by numerical evaluation of the respective difference equations.
The results will illustrate how the applied approximations affect the prediction of the steady-state.
Finally, a closed-form solution for the steady-state is derived in Sec.~\cref{sec:csa_ss_cont} and compared to real simulations of the CSA on the sphere.

\subsection{Introduction}
\label{sec:csa_intro}
For the subsequent investigation, three commonly chosen implementations of the CSA-ES in Alg.~\cref{alg:csa} will be tested.
For all CSA-variants, the cumulation path $\vb{s}$ of the $\sigma$-adaptation is given in terms of cumulation constant $\cs$ and recombined mutation direction $\rec{\vb{z}}$
\begin{align}\label{eq:csa_s}
	\gvec{s}{g+1} &= (1-\cs)\gvec{s}{g} + \sqrt{\cs(2-\cs)\mu} \rec{\vb{z}}^{(g+1)}. 
\end{align}
Then, the path length $\norm{\vb{s}^{(g+1)}}$ is measured and compared to its expected result under random selection.
Additional damping of the change is introduced via $D$ and $d_\sigma$, respectively.
One update rule for the $\sigma$-change is chosen as \cite{Han98} 
\begin{align}\begin{split}
		\label{eq:new_han_v1}
		\sigma^{(g+1)} = \sigma^{(g)}\EXP{\frac{1}{D}\qty(\norm{\vb{s}^{(g+1)}}/E_\chi-1)}.
\end{split}\end{align}
Alternatively,  $\sigma$ can also be updated via \cite[(44)]{hansen2023cma}
\begin{align}\begin{split}
		\label{eq:neq_han_v2}
		\sigma^{(g+1)} = \sigma^{(g)}\EXP{\frac{c_\sigma}{d_\sigma}\qty(\norm{\vb{s}^{(g+1)}}/E_\chi-1)},
\end{split}\end{align}
which is often chosen in newer implementation of the CSA.
$E_\chi$ is the expected value of a chi-distributed random variate $\chi \sim \norm{\mathcal{N}(\vb{0},\vb{1})}$ and one uses an approximation for large $N$
\begin{align}\begin{split}\label{eq:new_han_Echi} 		
		E_\chi\simeq\sqrt{N}\qty(1-1/4N+1/21N^2).
\end{split}\end{align} 
The CSA variants under investigation are parameterized as
\begin{subequations}
	\begin{align}
		&\text{Eq.~\cref{eq:new_han_v1}}\quad \text{with} \quad c_\sigma = 1/\sqrt{N},\quad D = c_\sigma^{-1}. \label{eq:sqrtN} \\
		&\text{Eq. \cref{eq:new_han_v1}}\quad \text{with} \quad c_\sigma = 1/N,\quad D = c_\sigma^{-1} \label{eq:linN}\\
		\begin{split}
			&\text{Eq. \cref{eq:neq_han_v2}}\quad \text{with}\quad c_\sigma = \frac{\mu+2}{N+\mu+5},\qq{and} \\
			&d_\sigma = 1+c_\sigma+2\max\qty(0,\sqrt{(\mu-1)/(N+1)}-1). \label{eq:han}
		\end{split}
	\end{align} 
\end{subequations}%
CSA implementations \cref{eq:sqrtN} and \cref{eq:linN} were investigated in more detail in \cite{HO01,Han98}.
\cite{Han98} derives the inverse proportionality $D = c_\sigma^{-1}$ with $c_\sigma = N^{-a}$ for $\frac{1}{2}\leq a \leq 1$ based on theoretical and experimental investigations on the sphere ($\mu\ll N$).
CSA \cref{eq:han} is a newer implementation that is part of the default CMA-ES, see also \cite{hansen2023cma} ($\mu_\mathrm{eff}$ due to weighted recombination was replaced by $\mu$).
The subsequent analysis will show that the three CSA variants have distinct adaptation properties on the sphere as a function of $\mu$ and $N$.
Furthermore, $c_\sigma$ and $D$ from \cref{eq:sqrtN} are re-derived using a steady-state analysis on the sphere by assuming $\mu \gg N$.

As a first step, $c_\sigma$ and $d_\sigma$ of \cref{eq:han} are further analyzed under the assumption $\mu\gg N$ and $\cs/d_\sigma$ is brought into a similar form as \cref{eq:new_han_v1}.
$c_\sigma$ yields simply
\begin{align}\begin{split}\label{eq:csigma_han}
		c_\sigma &= \frac{\mu(1+2/\mu)}{\mu(1+N/\mu+5/\mu)} \overset{\mu\rightarrow\infty}{\simeq} 1.
\end{split}\end{align}
The cumulation time parameter $c_\sigma$ approaches ``1" as $\mu$ is increased, which results in a faster cumulation in \cref{eq:csa_s}.
For the evaluation of damping $d_\sigma$, we introduce $d_\sigma = 1+c_\sigma+g(N,\mu)$ with $g(N,\mu) \coloneqq 2\max\qty(0,\sqrt{\frac{\mu-1}{N+1}}-1)$.
Now, $d_\sigma$ is inserted into the exponential of \cref{eq:neq_han_v2}, which yields 
%
\begin{math}		\EXP{\frac{c_\sigma\qty(\norm{\vb{s}^{(g+1)}}/E_\chi-1)}{1+c_\sigma+g(N,\mu)}} = \EXP{\frac{\norm{\vb{s}^{(g+1)}}/E_\chi-1}{1+1/c_\sigma+g(N,\mu)/c_\sigma}}.
\end{math}	
%
Comparing the last exponential with \cref{eq:new_han_v1}, one can derive the resulting damping parameter as $D = 1+1/c_\sigma+g(N,\mu)/c_\sigma$.
Note that the proportionality $d_\sigma\propto c_\sigma^{-1}$ holds.
Further analysis of the term $g(N,\mu)/c_\sigma$ yields for large $\mu \gg N$
\begin{align}\begin{split}\label{eq:neq_han_v2_check3}
		&\frac{g(N,\mu)}{c_\sigma} = 2\max\qty(0,\sqrt{\frac{\mu-1}{N+1}}-1)\frac{N+\mu+5}{\mu+2} \\
		&\simeq 2\qty(\sqrt{\frac{\mu-1}{N+1}}-1)\frac{N+\mu+5}{\mu+2} \qq{(large $\mu$)} \\
		&\simeq 2\sqrt{\mu/N} \qq{($\mu\rightarrow\infty$)}.
\end{split}\end{align}
For the last line of \cref{eq:neq_han_v2_check3}, the ``$-1$" after the square root was neglected and $N+1\simeq N$ and $\mu-1\simeq \mu$ were applied.
The resulting damping $D$ of \cref{eq:han} scales with $\sqrt{\mu}$ according to \cref{eq:neq_han_v2_check3} for fixed $N$ and $\cs$ approaches one with \cref{eq:csigma_han}.
Hence, CSA \cref{eq:han} employs a $\mu$-dependent damping which is in contrast to CSA \cref{eq:sqrtN} and CSA \cref{eq:linN}.
If $\mu$ is increased during active population control, the corresponding CSA damping is affected.
This will be investigated further.

\subsection{Steady-State Analysis}
\label{sec:csa_stst}

The steady-state analysis to be presented is related to \cite[p.~68]{Arn02}, where the CSA was studied on the sphere function under the assumption $\mu\ll N$ and $N\rightarrow \infty$.
Here, a different derivation path will be chosen by assuming large populations $\mu\gg N$.
The analysis to be conducted requires the evaluation of the cumulation path \cref{eq:csa_s} and the $\sigma$-update rule \cref{eq:new_han_v1}.
The approach will show that closed-form solutions for the CSA-dynamics can be obtained under certain approximations by assuming a steady-state on the sphere function.
Steady-state conditions on the sphere emerge due to its scale-invariance and $\sigma$ being reduced in accordance with the residual distance $R$, see \cref{eq:sign}.
As an example, $\sigma^*$ is constant in expectation in Fig.~\cref{fig:dyn_phi_csa_sa} on the left.
Imposing steady-state conditions, the generation-dependency of the evolution equations will vanish.
This will enable an analytic investigation of CSA-scaling properties as a function of cumulation and damping parameters.
Starting with the cumulation path update, one has
\begin{align}
	\gvec{s}{g+1} &= (1-\cs)\gvec{s}{g} + \sqrt{\cs(2-\cs)\mu} \gvec{\rec{z}}{g+1}\label{eq:csa_ss_1}  \\ 
	\begin{split}
		\snorm{g+1}^2 &= (1-\cs)^2\snorm{g}^2 
		+ 2(1-\cs)\sqrt{\cs(2-\cs)\mu}\sum_{i=1}^N s^{(g)}_i \rec{z_i}^{(g+1)} \\
		&\qquad+ \cs(2-\cs)\mu \znorm{g+1}^2 .
	\end{split}
	\label{eq:csa_ss_2}
\end{align}
One can decompose $\vb{s}$ and $\rec{\vb{z}}$ into a radial component (denoted by $A$, along unit vector $\vb{e}_A$) and a lateral $N-1$ dimensional vector $\vb{e}_B$ (denoted by $B$) perpendicular to $\vb{e}_A$, such that $\vb{s} = \vb{s}_A + \vb{s}_B$ and $\rec{\vb{z}} = \rec{\vb{z}_A} + \rec{\vb{z}_B}$ holds.
One can assume $\vb{e}_A = [1,0,...,0]$, such that $\rec{\vb{z}_A} = [z_A, 0, ..., 0]$ and $\rec{\vb{z}_B} = [0, z_2, ..., z_N]$.
Furthermore, $\vb{e}_A$ is defined to be positive pointing towards the optimizer.
The components along $\vb{e}_B$ are selectively neutral
and one has $\EV{s_i\rec{z_i}}=s_i\EV{\rec{z_i}}=0$ for $i \neq A$ and $\EV{s_i\rec{z_i}}=s_A \EV{\rec{\vb{z}_A}}$ for $i=A$.
Calculating the expectation of \cref{eq:csa_ss_2} yields
\begin{align}\begin{split}\label{eq:csa_ss_3}
		\EV{\snorm{g+1}^2} &= (1-\cs)^2\snorm{g}^2 
		+ 2(1-\cs)\sqrt{\cs(2-\cs)\mu}s^{(g)}_A \EV{\rec{z_A}^{(g+1)}} \\
		&\qquad+ \cs(2-\cs)\mu \EV{\znorm{g+1}^2}.
\end{split}\end{align}
Equation~\cref{eq:csa_ss_3} shows that an expression for the selected component $s_A$ of the cumulation path is necessary to model the CSA.
One has
\begin{align}\begin{split}\label{eq:csa_ss_4}
		s_A^{(g+1)} &= \gvec{s}{g+1}\gvec{e}{g+1}_A.
\end{split}\end{align}	
An expression for the unit vector $\gvec{e}{g+1}_A$ is required as a function of the applied mutation strength $\sigma$ and residual distance $R$.
The optimizer of the sphere is denoted by $\hat{\vb{y}}$ and the positional update
using recombinant $\rec{\vb{z}}$ is given by $\gvec{y}{g+1}=\gvec{y}{g}+\sigma^{(g)}\gvec{\rec{z}}{g+1}$, see also Alg.~\cref{alg:csa}.
One has
\begin{align}\begin{split}\label{eq:csa_ss_5}	
		R^{(g+1)}\gvec{e}{g+1}_A &= \hat{\vb{y}}-\gvec{y}{g+1} \\
		&= \hat{\vb{y}}-\qty(\gvec{y}{g}+\sigma^{(g)}\gvec{\rec{z}}{g+1}) \\
		&= \hat{\vb{y}} - \gvec{y}{g} - \sigma^{(g)}\gvec{\rec{z}}{g+1} \\
		&= R^{(g)}\gvec{e}{g}_A - \sigma^{(g)}\gvec{\rec{z}}{g+1} \\
		&= R^{(g)}\gvec{e}{g}_A - \frac{\sigma^{*,(g)}R^{(g)}}{N}\gvec{\rec{z}}{g+1}.
\end{split}\end{align}
Note that $\sigma=\sigma^* N / R$ from \cref{eq:sign} was introduced in the last line since the sphere steady-state is investigated.
Solving for the unit vector at $g+1$ yields
\begin{align}\begin{split}\label{eq:csa_ss_6}	
		\gvec{e}{g+1}_A = \frac{R^{(g)}}{R^{(g+1)}}\qty(\gvec{e}{g}_A-\frac{\sigma^{*,(g)}}{N}\gvec{\rec{z}}{g+1}).
\end{split}\end{align}
Now one can insert \cref{eq:csa_ss_1} and \cref{eq:csa_ss_6} into \cref{eq:csa_ss_4}, which yields
\begin{align}\begin{split}\label{eq:csa_ss_7}
		s_A^{(g+1)} &= \gvec{s}{g+1}\gvec{e}{g+1}_A  \\
		&= \qty[(1-\cs)\gvec{s}{g} + \sqrt{\cs(2-\cs)\mu} \gvec{\rec{z}}{g+1}]\frac{R^{(g)}}{R^{(g+1)}}\qty(\gvec{e}{g}_A-\frac{\sigma^{*,(g)}}{N}\gvec{\rec{z}}{g+1}) \\
		&= \frac{R^{(g)}}{R^{(g+1)}}
		\Big\{
		(1-\cs)s_A^{(g)} - (1-\cs)\frac{\sigma^{*,(g)}}{N}\sum_i s_i^{(g)}\rec{z_i}^{(g+1)} \\
		&+ \sqrt{\cs(2-\cs)\mu} z_A^{(g+1)}
		- \sqrt{\cs(2-\cs)\mu} \frac{\sigma^{*,(g)}}{N}\znorm{g+1}^2
		\Big\}.
\end{split}\end{align}	 
In the general case, the two factors $\frac{R^{(g)}}{R^{(g+1)}}$ and $\{\dots\}$ in \cref{eq:csa_ss_7} are dependent on each other since they both depend on the selected $z_A$-components.
Selection reduces the residual distance accordingly.
The goal is to evaluate $\EV{s_A^{(g+1)}}$.
To this end, the covariance between $\frac{R^{(g)}}{R^{(g+1)}}$ and $\{\dots\}$ in \cref{eq:csa_ss_7} will be neglected.
Experiments later will show that this assumption is not critical.
For the terms in $\{\dots\}$, taking the expectation yields $\EV{\sum_i s_i z_i} = s_A\EV{\rec{z_A}}$ with $\EV{z_i}=0$ for $i \neq A$.
One gets
\begin{align}\begin{split}\label{eq:csa_ss_7b}	
		\EV{s_A^{(g+1)}} &= 
		\EV{\frac{R^{(g)}}{R^{(g+1)}}}\Big\{(1-\cs)s_A^{(g)} -	(1-\cs) \frac{\sigma^{*,(g)}}{N} s_A^{(g)}\EV{\rec{z_A}^{(g+1)}} \\
		&+ \sqrt{\cs(2-\cs)\mu} \EV{\rec{z_A}^{(g+1)}}
		\sqrt{\cs(2-\cs)\mu}\frac{\sigma^{*,(g)}}{N}\EV{\znorm{g+1}^2}\Big\}.
\end{split}\end{align}	
The ratio of the residual distances can be expressed in terms of the sphere progress rate $\varphi^*$ using \cref{sec:dyn_phidef} and \cref{eq:signzero_approx_gam_v2} according to
\begin{align}\begin{split}\label{eq:csa_ss_8}	
		\EV{\frac{R^{(g)}}{R^{(g+1)}}} = \frac{1}{1 -\varphi^*/N}.
	\end{split}
\end{align}
Closed-form solutions of the CSA steady-state will require approximating $\EV{\frac{R^{(g)}}{R^{(g+1)}}} = 1$. 
This can be justified for $\varphi^*/N\ll1$, i.e., by comparably slow progress in relation to $N$.
In this case, the change of the residual distance between two generations is relatively small.
The effect of this approximation (and subsequent assumptions) will be investigated in more detail in Sec.~\cref{sec:csa_iteration}.
For the evaluation of \cref{eq:csa_ss_7b}, \cref{eq:csa_ss_8} is simplified by setting $\EV{\frac{R^{(g)}}{R^{(g+1)}}} = 1$, which yields
\begin{align}\begin{split}\label{eq:csa_ss_9}	
		\EV{s_A^{(g+1)}} &= 
		(1-\cs)s_A^{(g)} -	(1-\cs) \frac{\sigma^{*,(g)}}{N} s_A^{(g)}\EV{\rec{z_A}^{(g+1)}} \\
		&+ \sqrt{\cs(2-\cs)\mu} \EV{\rec{z_A}^{(g+1)}} \\
		&-\sqrt{\cs(2-\cs)\mu}\frac{\sigma^{*,(g)}}{N}\EV{\znorm{g+1}^2}.
\end{split}\end{align}	 
Now, the steady-state condition is imposed on \cref{eq:csa_ss_9} by setting $\EV{s_A^{(g+1)}} = s_A^{(g)} = s_A$.
Furthermore, the steady-state is assumed to hold for the selection of $\rec{\vb{z}}$-components.
Hence, the generation counter is dropped.
One has
\begin{align}\begin{split}\label{eq:csa_ss_10}	
		&s_A = 
		(1-\cs)s_A -	(1-\cs) \frac{\sigma^{*}}{N} s_A\EV{\rec{z_A}} \\
		&\qquad+ \sqrt{\cs(2-\cs)\mu} \EV{\rec{z_A}}
		-\sqrt{\cs(2-\cs)\mu}\frac{\sigma^{*}}{N}\EV{||\rec{\vb{z}}||^2} \\
		&s_A\qty(\cs+(1-\cs)\frac{\sigma^{*}}{N} \EV{\rec{z_A}}) = \sqrt{\cs(2-\cs)\mu} \qty(\EV{\rec{z_A}}-\frac{\sigma^{*}}{N}\EV{||\rec{\vb{z}}||^2}),
\end{split}\end{align}	 
such that solving for $s_A$ yields
\begin{align}\begin{split}\label{eq:csa_ss_11}	
		s_A &= \frac{\sqrt{\cs(2-\cs)\mu} \qty(\EV{\rec{z_A}}-\frac{\sigma^{*}}{N}\EV{||\rec{\vb{z}}||^2})}
		{\cs+(1-\cs)\frac{\sigma^{*}}{N} \EV{\rec{z_A}}}.
\end{split}\end{align}	 
In \cite{Arn02}, the second term in the denominator is neglected by assuming $\frac{\sigma^{*}\EV{\rec{z_A}}}{N} \ll1$.
This can be justified for large $N$ and small $\sigma^*$ and by assuming small progress contribution of $\EV{\rec{z_A}}$.
Note that demanding small $\sigma^*/N$ implicitly contains the condition that $\mu\ll N$ as small $\mu$ yield comparably small $\sigma^*$-levels.
Here, a different approach is taken by assuming $\mu\gg\sqrt{N}$, which yields large mutation strengths $\sigma^*$.
Analogous to $s_A$ in \cref{eq:csa_ss_10}, the steady-state condition is imposed on the squared norm $\EV{\snorm{g+1}^2} = \snorm{g}^2 = ||\vb{s}||^2$ in \cref{eq:csa_ss_3}, which gives
\begin{align}\begin{split}\label{eq:csa_ss_12}
		\cs(2-\cs)||\vb{s}||^2 &= 
		2(1-\cs)\sqrt{\cs(2-\cs)\mu}s_A\EV{\rec{z_A}} + \cs(2-\cs)\mu\EV{||\rec{\vb{z}}||^2} \\
		||\vb{s}||^2 &= \frac{2(1-\cs)\sqrt{\cs(2-\cs)\mu}}{\cs(2-\cs)}s_A\EV{\rec{z_A}} + \mu\EV{||\rec{\vb{z}}||^2}
\end{split}\end{align}
In the steady-state, \cref{eq:csa_ss_11} can be inserted into \cref{eq:csa_ss_12}, which yields
\begin{align}\begin{split}\label{eq:csa_ss_13}
		||\vb{s}||^2 &= \mu\EV{||\rec{\vb{z}}||^2} + 2(1-\cs)\mu
		\frac{\EV{\rec{z_A}}^2 - \frac{\sigma^{*}}{N}\EV{\rec{z_A}}\EV{||\rec{\vb{z}}||^2}}
		{\cs+(1-\cs)\frac{\sigma^{*}}{N}\EV{\rec{z_A}}}.
\end{split}\end{align}
A second condition is required to obtain an expression for the steady-state $||\vb{s}||^2$.
To this end, one analyzes the update rule \cref{eq:new_han_v1}
\begin{align}\begin{split}
		\label{eq:csa_ss_14}
		\sigma^{(g+1)} = \sigma^{(g)}\EXP{\frac{1}{D}\qty(\frac{\norm{\vb{s}^{(g+1)}}}{E_\chi}-1)}.
\end{split}\end{align}
On the sphere, one uses $\sigma^{(g)} = \sigma^{*,(g)}R^{(g)}/N$, such that
\begin{align}\begin{split}
		\label{eq:csa_ss_15}
		\sigma^{*,(g+1)} = \sigma^{*,(g)}\frac{R^{(g)}}{R^{(g+1)}}\EXP{\frac{1}{D}\qty(\frac{\norm{\vb{s}^{(g+1)}}}{E_\chi}-1)}.
\end{split}\end{align}

To provide closed-form solutions of \cref{eq:csa_ss_13} by using \cref{eq:csa_ss_15}, further approximations are necessary.
To this end, an expression for the ratio $\frac{R^{(g)}}{R^{(g+1)}}$ is needed.
The positional change in search space is given by $\Delta_\varphi = R^{(g)}-R^{(g+1)}$ (see \cref{sec:dyn_phidef}).
The ratio $\frac{R^{(g)}}{R^{(g+1)}}$ in terms of normalized $\Delta^*_\varphi$ (see \cref{eq:csa_ss_8}) can be obtained from 
\begin{align}\begin{split}
		\label{eq:csa_ss_15b}
		\frac{R^{(g)}}{R^{(g+1)}} &= \frac{1}{1 - \Delta^*_\varphi/N}.
\end{split}\end{align}
The calculation of $\EV{\frac{R^{(g)}}{R^{(g+1)}}\EXP{\frac{1}{D}\qty(\frac{\norm{\vb{s}^{(g+1)}}}{E_\chi}-1)}}$ in \cref{eq:csa_ss_15} is unfeasible due to covariance between $R^{(g+1)}$ and $\norm{\vb{s}^{(g+1)}}$, and $\norm{\vb{s}^{(g+1)}}$ being within the exponential function.
Since a steady-state analysis is performed, the random variate $\frac{R^{(g)}}{R^{(g+1)}}$ is replaced by its expected value \cref{eq:csa_ss_8}, for which the analytic expression is known.
Hence, one sets in \cref{eq:csa_ss_15}
\begin{align}\begin{split}\label{eq:csa_ss_16}	
	\frac{R^{(g)}}{R^{(g+1)}} = \EV{\frac{R^{(g)}}{R^{(g+1)}}} = \frac{1}{1 - \varphi^*/N} = 1 + \frac{\varphi^*}{N} + O\qty(\frac{\varphi^{*2}}{N^2}),
\end{split}\end{align}
with Taylor-expansion applied to obtain the last term.
Similarly, the steady-state condition $\norm{\vb{s}^{(g+1)}} = \EV{\snorm{g+1}} = ||\vb{s}||$ is applied in \cref{eq:csa_ss_15}.
Then, the exponential function of \cref{eq:csa_ss_15} is expanded as

\begin{align}\begin{split}
		\label{eq:csa_ss_17}
		\EXP{\frac{1}{D}\qty(\frac{\norm{\vb{s}^{(g+1)}}}{E_\chi}-1)} = \EXP{\frac{1}{D}\qty(\frac{||\vb{s}||}{E_\chi}-1)} &= 1+\frac{||\vb{s}||-E_\chi}{D E_\chi} + O\qty(\qty[\frac{||\vb{s}||-E_\chi}{D E_\chi}]^2).
\end{split}\end{align}
In \cref{eq:csa_ss_16} and \cref{eq:csa_ss_17}, the respective higher order terms of the Taylor-expansions must be dropped for closed-form solutions.
The term $\varphi^{*2}/ N^2 \ll 1$ is neglected assuming slow progress rates (see also discussion of \cref{eq:csa_ss_8}).
For \cref{eq:csa_ss_17} is should be noted that $||\vb{s}||$ scales approximately with $\sqrt{N}$ for random selection.
Furthermore, $E_\chi\simeq\sqrt{N}$ with $O(1/\sqrt{N})$ in \cref{eq:new_han_Echi}.
Hence, $O\qty(\qty[(||\vb{s}||-E_\chi)/D E_\chi]^2)$ is neglected by assuming sufficiently high damping $D$.
The evaluation of the product of the remaining terms, which is necessary for \cref{eq:csa_ss_15}, yields
\begin{align}\begin{split}\label{eq:csa_ss_17b}
		\qty[1 + \frac{\varphi^*}{N}]\qty[1+\frac{||\vb{s}||-E_\chi}{D E_\chi}] = 1 + \frac{\varphi^*}{N} +\frac{||\vb{s}||-E_\chi}{D E_\chi} + O\qty(\frac{\varphi^{*}}{DN}).
\end{split}\end{align}
As before, the higher order term $O\qty(\frac{\varphi^{*}}{DN})$ must be neglected by assuming $\varphi^*\ll N$ and sufficiently high damping.
One obtains an approximation of the update rule \cref{eq:csa_ss_15} as
\begin{align}\begin{split}\label{eq:csa_ss_18b}
		\sigma^{*,(g+1)} &= \sigma^{*,(g)}
		\qty[1 + \frac{\varphi^*}{N} +\frac{||\vb{s}||-E_\chi}{D E_\chi}].
\end{split}\end{align}
Within the steady-state on the sphere, it holds for the normalized mutation strength
\begin{align}\begin{split}\label{eq:csa_ss_18c}
		\EV{\sigma^{*,(g+1)}} &= \sigma^{*,(g)}= \sigma^{*}.
\end{split}\end{align}
Since \cref{eq:csa_ss_18c} holds, the terms in $[\dots]$ of \cref{eq:csa_ss_18b} must be equal to one in the steady-state, such that
\begin{align}\begin{split}
		\label{eq:csa_ss_19}
		\frac{\norm{\vb{s}}-E_\chi}{D E_\chi} &= -\frac{\varphi^*}{N} \\
		||\vb{s}|| &= E_\chi\qty[1-\frac{\varphi^* D}{N}] \\
		||\vb{s}||^2 &= E_\chi^2\qty[1-\frac{2\varphi^* D}{N}+\frac{\varphi^{*2} D^2}{N^2}].
\end{split}\end{align}
For $E_\chi$, one uses the approximation $E_\chi\simeq\sqrt{N}\qty(1-\frac{1}{4N}+\frac{1}{21N^2})$ from \cref{eq:new_han_Echi}, see also \cite{hansen2023cma}, such that after squaring one gets
\begin{align}\begin{split}
		\label{eq:csa_ss_20}
		E_\chi^2 &= N\qty(1+O\qty(1/N)).
\end{split}\end{align}
In principle, \cref{eq:csa_ss_19} needs to be inserted into \cref{eq:csa_ss_13}.
However, two approximations are applied first to omit the calculation of the squared progress rate $\varphi^{*2}$ \footnote{
	If the CSA \cite[p.~16]{Arn02} is used, 
	the $\sigma$-update rule yields $\sigma^{(g+1)} = \sigma^{(g)}\EXP{\frac{\norm{\vb{s}^{(g+1)}}^2-E_{\chi^2}}{2 D E_{\chi^2}}}$,
	which directly evaluates the squared norm and $E_{\chi^2} = N$ for the expectation of a chi-squared distributed variate.
	In this case, approximation \cref{eq:csa_ss_20} is not necessary and there is no higher order term $\frac{\varphi^{*2} D^2}{N^2}$ as in \cref{eq:csa_ss_19} emerging when squaring $||\vb{s}||$.
	By neglecting higher order terms of the Taylor-expansion, as it was done in \cref{eq:csa_ss_18b}, one obtains $||\vb{s}||^2 = N - 2D\varphi^*$ as in \cref{eq:csa_ss_19b}.
	Theoretical results and corresponding experiments show slightly better agreement when using the CSA from \cite{Arn02}.}.
In \cref{eq:csa_ss_20}, the higher order terms $O(1/N)$ are neglected for large $N$.
Furthermore, the higher order term $\frac{\varphi^{*2} D^2}{N^2}$ in \cref{eq:csa_ss_19} is dropped by assuming $\varphi^{*2} D^2/N^2\ll1$.
Hence, \cref{eq:csa_ss_19} is simplified as
\begin{align}\begin{split}\label{eq:csa_ss_19b}
		||\vb{s}||^2 &\simeq E^2_{\chi}\qty[1-\frac{2\varphi^* D}{N}] \simeq N - 2D\varphi^*.
\end{split}\end{align}
Later, the different applied approximations are evaluated by means of iterating the steady-state CSA-dynamics.
To this end, the approximation \cref{eq:csa_ss_19b} is iterated within the dynamics of $\sigma^{*,(g)}$.
Since $1+\frac{\norm{\vb{s}}^2-N}{2 D N}+\frac{\varphi^*}{N}=1$ from \cref{eq:csa_ss_19b}, the corresponding $\sigma^*$-update in terms of $||\vb{s}||^2$ is given as
\begin{align}\begin{split}\label{eq:csa_ss_20a}
		\sigma^{*,(g+1)} &= \sigma^{*,(g)}
		\qty[1 + \frac{\varphi^*}{N} +\frac{\norm{\vb{s}^{(g+1)}}^2-N}{2 D N}].
\end{split}\end{align}
Finally, \cref{eq:csa_ss_19b} is inserted into \cref{eq:csa_ss_13}. One obtains the CSA steady-state condition
\begin{align}\begin{split}\label{eq:csa_ss_20b}
		N - 2D\varphi^* &= \mu\EV{||\rec{\vb{z}}||^2}
		+ 2(1-\cs)\mu
		\frac{\EV{\rec{z_A}}^2 - \frac{\sigma^{*}}{N}\EV{\rec{z_A}}\EV{||\rec{\vb{z}}||^2}}
		{\cs+(1-\cs)\frac{\sigma^{*}}{N}\EV{\rec{z_A}}}.
\end{split}\end{align}
The result \cref{eq:csa_ss_20b} has no functional dependency on $||\vb{s}||^2$ and $s_A$ any more.
Instead, the steady-state is characterized by expected values of the selected mutation components (which includes the progress rate $\varphi^*$).
As shown below, they are only functions of $\sigma^*$, $N$, $\mu$, and $\vt$.
For given ES parameters and $N$, the only remaining parameter is the normalized mutation strength $\sigma^*$, which will be solved for in Sec.~\cref{sec:csa_ss_cont}.

The expected values $\varphi^*$, $\EV{\rec{z_A}}$, and $\EV{||\rec{\vb{z}}||^2}$ are known quantities.
Furthermore, there are different approximation orders thereof (more details are given in \cite{Arn02}).
Hence, two sets of results will be tested in the following.
The first set uses the highest approximation quality with
\begin{subequations}
	\begin{align}
		\varphi^* &\qq{from \cref{eq:sph_Ndep_pc}} \label{eq:csa_ss_full1} \\
		\EV{\rec{z_A}} &= \frac{\CTHETA}{\sqrt{1+\sigma^{*2}/2N}},\qq{from \cite[(5.7)]{Arn02}} \label{eq:csa_ss_full2} \\
		\EV{||\rec{\vb{z}}||^2} &= 
		\frac{1}{\mu}\qty(N + \frac{e^{1,1}_\vt + (\mu-1)e^{2,0}_\vt}{1+\sigma^{*2}/2N} - \frac{N-1}{N}\frac{\sigma^*\CTHETA}{\sqrt{1+\sigma^{*2}/2N}}),\qq{from \cite[(5.11)]{Arn02}.} \label{eq:csa_ss_full3}
	\end{align}
\end{subequations}
For the second set, the large population approximation is applied to all quantities.
This set will enable closed-form solutions of \cref{eq:csa_ss_20b} for the scaling properties of the CSA.
One has
\begin{subequations}
	\begin{align}
		\varphi^* &= \sqrt{2N}\CTHETA - \sigma^{*2}/2\mu \qq{from \cref{sec:dyn_phi_large}} \label{eq:csa_ss_large1} \\
		\EV{\rec{z_A}} &= \sqrt{2N}\CTHETA/\sigma^*, \qq{from \cref{eq:csa_ss_full2} with $\sqrt{1+\sigma^{*2}/2N} \simeq \sigma^*/\sqrt{2N}$} \label{eq:csa_ss_large2} \\
		\EV{||\rec{\vb{z}}||^2} &= N/\mu, \qq{from \cite[(5.2)]{Arn02}}. \label{eq:csa_ss_large3}
	\end{align}
\end{subequations}

\subsection{Iteration of CSA Steady-State Equations}
\label{sec:csa_iteration}

Before continuing the analysis of the steady-state condition \cref{eq:csa_ss_20b}, the effects of the approximations which were applied throughout Sec.~\cref{sec:csa_intro} are investigated first.
To this end, the dynamics of the CSA will be evaluated by means of evolution equations, i.e., using an iterative mapping $g\mapsto g+1$ of the relevant quantities $\snorm{g}^2$, $s_A^{(g)}$, and $\sigma^{*,(g)}$.
The analysis to be conducted is related (to some extent) to the CSA-analysis on the ellipsoid model in \cite{BH14}.
However, in \cite{BH14} it was conducted in the limit $N\rightarrow\infty$ ($\mu\ll N$).
Here, the analysis will focus on the steady-state dynamics of the CSA in the limit $\mu\gg N$ on the sphere function.
Furthermore, the investigation will only consider the dynamics \emph{in expectation} by neglecting fluctuation of the underlying quantities.
The advantage of iterating the respective dynamics is that one can test various approximation stages comparably easy since no closed-form solutions are needed.
Hence, one can apply the formulae from \cref{eq:csa_ss_full1},\cref{eq:csa_ss_full1}, and \cref{eq:csa_ss_full3} and determine its predicted steady-state, even though closed-form solutions are not available.
As already mentioned, closed-form solutions will be obtained in Sec.~\cref{sec:csa_ss_cont} by using \cref{eq:csa_ss_large1}, \cref{eq:csa_ss_large2}, and \cref{eq:csa_ss_large3}.

To illustrate the effects of certain approximations, the subsequent \emph{four iteration schemes} will be tested.
All iterations are initialized at $\snorm{0}^2=s_A^{(0)}=0$ and $\sigma^{*,(0)} = \signzero$ from \cref{eq:signzero_approx}.
For simplicity, the notation $\EV{(\cdot)^{(g+1)}}$ for the updated quantities on the LHS will be dropped.
All obtained values can be understood as values in expectation.
\begin{figure}[t]
	\centering
	\includegraphics[]{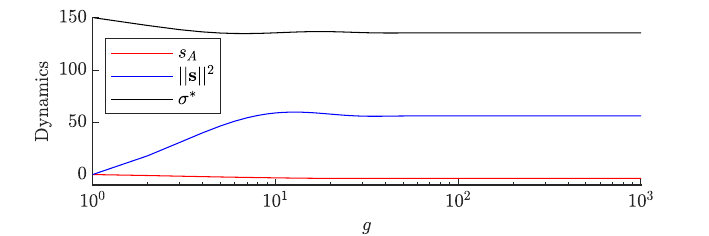}
	\caption{Iteration~\cref{eq:iter_csa_1a} of CSA-dynamics on the sphere $N=100$ for $\mu=1000$ ($\vt=1/2$).
		One measures $\signss\approx135.51$ with $\sigma^*_0\approx154.5$, giving $\gamma\approx0.88$.}
	\label{fig:iter_csa_dyn}
\end{figure}\noindent

Iteration 1A (with expected values \cref{eq:csa_ss_full1},\cref{eq:csa_ss_full2}, and \cref{eq:csa_ss_full3}) yields
\begin{align}\begin{split}\label{eq:iter_csa_1a}
		\snorm{g+1}^2 &= (1-\cs)^2\snorm{g}^2 
		+ 2(1-\cs)\sqrt{\cs(2-\cs)\mu}s^{(g)}_A \EV{\rec{z_A}^{(g+1)}}  \\
		&+ \cs(2-\cs)\mu\EV{\znorm{g+1}^2} 	\qq{from \cref{eq:csa_ss_3}.} \\				
		s_A^{(g+1)} &= 
		\qty[1-\frac{\varphi^*}{N}]^{-1}\Big\{(1-\cs)s_A^{(g)} -	(1-\cs) \frac{\sigma^{*,(g)}}{N} s_A^{(g)}\EV{\rec{z_A}^{(g+1)}} \\
		&+ \sqrt{\cs(2-\cs)\mu} \EV{\rec{z_A}^{(g+1)}}
		\sqrt{\cs(2-\cs)\mu}\frac{\sigma^{*,(g)}}{N}\EV{\znorm{g+1}^2}\Big\}	\qq{from \cref{eq:csa_ss_7b}.} \\
		\sigma^{*,(g+1)} &= \sigma^{*,(g)}\qty[1-\frac{\varphi^*}{N}]^{-1}\EXP{\frac{1}{D}\qty(\frac{\norm{\vb{s}^{(g+1)}}}{E_\chi}-1)} \qq{from \cref{eq:csa_ss_15}.}
\end{split}\end{align}

Iteration 1B (with expected values \cref{eq:csa_ss_full1},\cref{eq:csa_ss_full2}, and \cref{eq:csa_ss_full3}) yields
\begin{align}\begin{split}\label{eq:iter_csa_1b}
		\snorm{g+1}^2 &= \dots \qq{from Iter.~1A} \\				
		s_A^{(g+1)} &= (1-\cs)s_A^{(g)} -	(1-\cs) \frac{\sigma^{*,(g)}}{N} s_A^{(g)}\EV{\rec{z_A}^{(g+1)}} \\
		&+ \sqrt{\cs(2-\cs)\mu} \EV{\rec{z_A}^{(g+1)}}
		\sqrt{\cs(2-\cs)\mu}\frac{\sigma^{*,(g)}}{N}\EV{\znorm{g+1}^2} \qq{from \cref{eq:csa_ss_9}.} \\
		\sigma^{*,(g+1)} &= \dots \qq{from Iter.~1A}
\end{split}\end{align}

Iteration 2A (with large population approximations \cref{eq:csa_ss_large1}, \cref{eq:csa_ss_large2}, and \cref{eq:csa_ss_large3}) yields
\begin{align}\begin{split}\label{eq:iter_csa_2a}
		\snorm{g+1}^2 &= (1-\cs)^2\snorm{g}^2 
		+ 2(1-\cs)\sqrt{\cs(2-\cs)\mu}s^{(g)}_A \EV{\rec{z_A}^{(g+1)}}  \\
		&+ \cs(2-\cs)\mu\EV{\znorm{g+1}^2} 	\qq{from \cref{eq:csa_ss_3}.} \\	
		s_A^{(g+1)} &= (1-\cs)s_A^{(g)} -	(1-\cs) \frac{\sigma^{*,(g)}}{N} s_A^{(g)}\EV{\rec{z_A}^{(g+1)}} \\
		&+ \sqrt{\cs(2-\cs)\mu} \EV{\rec{z_A}^{(g+1)}}
		\sqrt{\cs(2-\cs)\mu}\frac{\sigma^{*,(g)}}{N}\EV{\znorm{g+1}^2} \qq{from \cref{eq:csa_ss_9}.} \\
		\sigma^{*,(g+1)} &= \sigma^{*,(g)}
		\qty[1 + \frac{\varphi^*}{N} +\frac{\norm{\vb{s}^{(g+1)}}-E_\chi}{D E_\chi}] \qq{from \cref{eq:csa_ss_18b}.}
\end{split}\end{align}

Iteration 2B (with large population approximations \cref{eq:csa_ss_large1}, \cref{eq:csa_ss_large2}, and \cref{eq:csa_ss_large3}) yields
\begin{align}\begin{split}\label{eq:iter_csa_2b}
		\snorm{g+1}^2 &= \dots \qq{from Iter.~2a} \\
		s_A^{(g+1)} &= \dots \qq{from Iter.~2a}\\
		\sigma^{*,(g+1)} &= \sigma^{*,(g)}
		\qty[1 + \frac{\varphi^*}{N} +\frac{\norm{\vb{s}^{(g+1)}}^2-N}{2 D N}] \qq{from \cref{eq:csa_ss_20a}.}
\end{split}\end{align}
Iteration 1A~\cref{eq:iter_csa_1a} and Iteration 1B~\cref{eq:iter_csa_1b} include the expected values of comparably high accuracy by including respective higher order terms.
Exemplary dynamics of the first iteration are shown in Fig.~\cref{fig:iter_csa_dyn}, showing that the three quantities reach their steady-state values relatively fast.
Iteration 1B is analogous to 1A, with the additional approximation $\frac{R^{(g)}}{R^{(g+1)}}=1$ applied to $s_A$.
Iteration 2A~\cref{eq:iter_csa_2a} and Iteration 2B~\cref{eq:iter_csa_2b} use the large population approximation of the underlying expected values.
Furthermore, $\frac{R^{(g)}}{R^{(g+1)}}=1$ for $s_A$ and the $\sigma^*$-update was Taylor-expanded, such that higher order terms were neglected.
As will be shown, Iteration 2B corresponds to the closed-form analytic solution which will be obtained in Sec.~\cref{sec:csa_ss_cont}.

\begin{figure}[t]
	\begin{subfigure}{0.5\textwidth}
		\centering
		\includegraphics[]{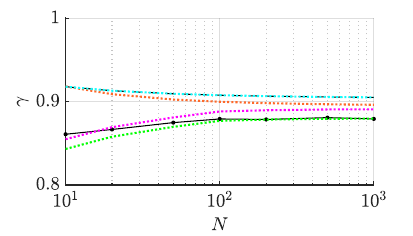}
		\caption{$\mu=1000$ with CSA~\cref{eq:new_han_v1}.}
		\label{fig:compare_iter1}
	\end{subfigure}
	\begin{subfigure}{0.5\textwidth}
		\centering
		\includegraphics[]{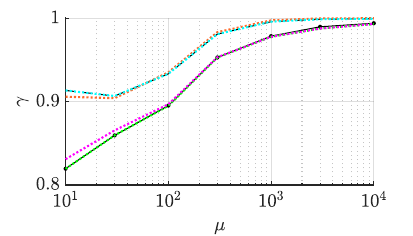}
		\caption{$N=100$ with CSA~\cref{eq:neq_han_v2}.}
		\label{fig:compare_iter2}
	\end{subfigure}
	\caption{Steady-state $\gamma$ for $\muilam{\mu}{\lambda}$-CSA-ES ($\vt=1/2$) with measured data (solid black with data points) and prediction \cref{eq:csa_gamma_b_v2} (dashed black).
		The dotted colored curves correspond to the Iterations 1A \cref{eq:iter_csa_1a} (green), 1B \cref{eq:iter_csa_1b} (magenta), 2A \cref{eq:iter_csa_2a} (orange), and 2B \cref{eq:iter_csa_2b} (cyan).
		Iteration 2B agrees with $\gamma$ from \cref{eq:csa_gamma_b_v2}, showing overlapping curves.}
	\label{fig:compare_iter}
\end{figure}

In Fig.~\cref{fig:compare_iter}, two parameter variations for the steady-state analysis of the CSA on the sphere are conducted.
To this end, the measured steady-state $\signss$ are determined.
Given the approach \cref{eq:signzero_approx_gam_v2}, $\signss$ is measured from experiments, $\sigma^*_0$ is determined numerically
\footnote{It is obtained from one-generation experiments for $N<100$ by evaluating $\varphi^{(g)} = R^{(g)}-\EV{R^{(g+1)}}$ and normalizing via $\varphi^* = \varphi N/R^{(g)}$ ($10^4$ trials).
	For $N\geq100$, it is obtained by numerical solving of \cref{eq:sph_Ndep_pc} since the effect $O\qty(N^{-1/2})$-terms becomes negligible.
}, and $\gamma= \signss/\sigma^*_0$ is evaluated as a reference (see data points).
Note that $\sigma^*_0$ is used since $\signzero$ introduces approximation errors, see Fig.~\cref{fig:phi_sign}.
For the iterations, one evaluates the steady-state $\signss$ as shown in Fig.~\cref{fig:iter_csa_dyn}.
Since the iterations use different progress rate formulae, the obtained $\signss$ is normalized by their respective progress rate zero (numerically obtained zero of \cref{eq:sph_Ndep_pc} for Iter.~1A and 1B, \cref{eq:signzero_approx} for Iter.~2A and 2B).
This distinction is important since it will largely explain the observed deviations.
In Fig.~\cref{fig:compare_iter}, Iteration~1A shows the best agreement with measured values.
The largest deviations occur at small $N=10$ due to missing higher order $N$-dependent terms of the progress rate \cref{eq:sph_Ndep_pc}.
At larger $N$-values the observed deviations become very small.
Iteration 1B introduces some notable deviations, however, they are comparably small.
By introducing the large population approximation (Iter.~2A and 2B),
larger deviations can be observed, especially for small $N$ and for $\mu\ll N$, which was expected based on the applied approximations.
These deviations are not due to the Taylor-expansion of the $\sigma^*$-update (using the full exponential \cref{eq:csa_ss_15} yields very similar results as Iter.~2A, not shown in plot), but can be attributed to the large population approximation.
As an example, the term $\EV{||\rec{\vb{z}}||^2}$ of \cref{eq:csa_ss_full3} contains additional $\mu$- and $N$-dependent terms which are not present in \cref{eq:csa_ss_large3}.
The same holds for progress rate $\varphi^*$ and $\EV{\rec{z_A}}$.
Unfortunately, closed-form solutions of the steady-state by including the higher order terms cannot be obtained.
However, one can still deduce important scaling properties of the CSA w.r.t.~$\mu$ and $N$ by using the large population approximation.
Furthermore, in the context of adaptive population control on noisy or multimodal functions, the exact prediction of $\gamma$ has not the highest priority.
More important is the scaling property of $\gamma$ w.r.t.\ $\mu$ and $N$ to understand how the $\sigma$-adaptation changes as population or dimensionality parameters change.
Hence, the analytic derivation is continued based on the large population approximation to investigate the scaling properties of the CSA-adaptation.

\subsection{Scaling Properties of the CSA}
\label{sec:csa_ss_cont}

Given the result~\cref{eq:csa_ss_20b}, the analytic derivation is now continued by inserting the respective expected values \cref{eq:csa_ss_large1}, \cref{eq:csa_ss_large2}, and \cref{eq:csa_ss_large3}.
For brevity, one introduces the following re-occurring factor
\begin{align}\begin{split}\label{eq:csa_ss_21}
		a &\coloneqq \sqrt{2N}\CTHETA.
\end{split}\end{align}
Then, one obtains the steady-state condition%
\begin{align}\begin{split}\label{eq:csa_ss_24}
		N - 2D\qty(a-\frac{\sigma^{*2}}{2\mu}) &= N + 2(1-\cs)\mu
		\frac{\frac{a^2}{\sigma^{*2}} - \frac{a}{\mu}}
		{\cs+(1-\cs)\frac{a}{N}} \\
		a-\frac{\sigma^{*2}}{2\mu} = -&\frac{1-\cs}{D(\cs+(1-\cs)a/N)}\frac{a\mu}{\sigma^{*2}}\qty(a-\frac{\sigma^{*2}}{\mu}) \\
		\frac{\sigma^{*2}}{a\mu}  \frac{1-\frac{\sigma^{*2}}{2a\mu}}{1-\frac{\sigma^{*2}}{a\mu}} &= -\frac{1-\cs}{D(\cs+(1-\cs)a/N)}.
\end{split}\end{align}
The following substitutions are introduced
\begin{align}\begin{split}\label{eq:csa_ss_25}
		x\coloneqq \frac{\sigma^{*2}}{a\mu},\quad b \coloneqq \frac{1-\cs}{D(\cs+(1-\cs)a/N)},
\end{split}\end{align}
such that \cref{eq:csa_ss_24} yields the second order equation
\begin{align}\begin{split}\label{eq:csa_ss_26}
		x^2 + 2(b-1)x - 2b = 0.
\end{split}\end{align}
The positive solution of \cref{eq:csa_ss_26} yields
\begin{align}\begin{split}\label{eq:csa_ss_27}
		x = \sqrt{1+b^2}-b+1.
\end{split}\end{align}
Note that $x>0$ for any $b>0$ $(0<\cs<1)$.
Now, $x$ from \cref{eq:csa_ss_25} is back-substituted into \cref{eq:csa_ss_27}.
For brevity, $b$ is not back-substituted.
Note that it is a function of the cumulation parameter, damping, and dimensionality.
In the end, one can solve for $\sigma^*=\signss$ and gets
\begin{align}\begin{split}\label{eq:csa_sign_res}
		\signss &= (2N)^{1/4}\sqrt{\CTHETA\mu}
		\qty[\sqrt{1+b^2}
		-b
		+ 1
		]^{1/2}.
\end{split}\end{align}
In principle, \cref{eq:csa_sign_res} can be used to predict (within the applied approximations) the steady-state $\signss$ as a function of the CSA parameters.
For our application, it is more useful to look at the ratio $\gamma = \signss/\signzero$, i.e., at $\signss$ relative to the second zero \cref{eq:signzero_approx_gam_v2}.
This has another advantage.
By equating \cref{eq:signzero_approx_gam_v2} with \cref{eq:csa_sign_res}, one can investigate the parameter dependency of the CSA as a function of $\gamma$.
Inserting $\signss=\gamma\signzero$ from \cref{eq:signzero_approx_gam_v2} and solving for $\gamma$, one gets
\begin{equation}\label{eq:csa_gamma_b_v2}
	\gamma = \sqrt{\frac12\qty(\sqrt{1+b^2}-b + 1)}.
\end{equation}
The result \cref{eq:csa_gamma_b_v2} predicts the steady-state $\gamma$ as a function of the CSA-parameters.
Assuming a constant $\gamma$, one can also solve for $b$ by assuming $1/\sqrt{2}<\gamma<1$ ($\signss$ close to $\signzero$).
Note that one has $b=\frac{1-\cs}{D(\cs+(1-\cs)a/N)}=\bcoeff$.
The solution for $b$ yields
\begin{equation}\label{eq:csa_b_gamma}
	b = \bcoeff =  \frac{2(\gamma^2-\gamma^4)}{2\gamma^2-1}. 
\end{equation}
The result \cref{eq:csa_b_gamma} is interesting as it relates the cumulation parameters and the dimensionality $N$ to a term on the right-hand side which only depends on a scale factor $\gamma$.
Note that the $\mu$-dependent prefactor has been canceled.
The limit $D\rightarrow\infty$ ($0<\cs<1$) yields $b\rightarrow0$ and $\gamma\rightarrow1$.
In this limit, $\varphi^*\rightarrow0$ by approaching its second zero.
Demanding the right side of \cref{eq:csa_b_gamma} be independent of $\mu$ and $N$ (with $O(1/\sqrt{N})$), one gets
\begin{align}\begin{split}\label{eq:csa_ss_31_main}
		D = c_1\sqrt{N}\quad (\text{constant } c_1>0).
\end{split}\end{align}
For $\cs D$ to be independent of $N$, one demands 
\begin{align}\begin{split}\label{eq:csa_ss_32_main}
		\cs = c_2D^{-1}\quad (\text{constant } c_2>0).
\end{split}\end{align}
Inserting \cref{eq:csa_ss_31_main} and \cref{eq:csa_ss_32_main} into \cref{eq:csa_b_gamma}, one gets
\begin{align}\begin{split}\label{eq:csa_b}
		b = \frac{1}{c_2/(1-O(1/\sqrt{N})) + \sqrt{2}c_1\CTHETA} = \frac{2(\gamma^2-\gamma^4)}{2\gamma^2-1}.
\end{split}\end{align}
As an example, one may set $c_1=c_2=1$ in \cref{eq:csa_b}.
This choice agrees with the cumulation parameter recommendation of \cite[p.~12]{Han98}, which was used in \cref{eq:sqrtN}.
For sufficiently large $N$, one may neglect $O(1/\sqrt{N})$ and \cref{eq:csa_b} yields
\begin{align}\begin{split}\label{eq:csa_ss_34_main}
		b \simeq 1/(1+\sqrt{2}\CTHETA).
\end{split}\end{align}
By inserting \cref{eq:csa_ss_34_main} back into \cref{eq:csa_gamma_b_v2}, one gets with $\CTHETA=\CTHETAlong$ (see \cite{OB23}, $\Phi^{-1}$ denoting the quantile function of the normal distribution)
\begin{align}\begin{split}\label{eq:csa_ss_35}
		\gamma \approx 0.90 & \text{ for $\vt=1/2$}\quad
		(\gamma \approx0.92\text{ for $\vt=1/4$}).
\end{split}\end{align}
A few important results can be deduced from the analysis.
Only CSA~\cref{eq:sqrtN} maintains a constant $\gamma$ independent of $N$ or $\mu$.
CSA \cref{eq:linN} yields $b\rightarrow0$ and $\gamma\rightarrow1$ for increasing $N$ in \cref{eq:csa_b_gamma}.
Hence, it operates increasingly closer to the second zero with increasing $N$.
For CSA~\cref{eq:han}, one has $\cs\simeq1$ and the respective damping $D$ increasing as $\sqrt{\mu/N}$ via \cref{eq:neq_han_v2_check3}.
In this case, one also has $b\rightarrow0$ and $\gamma\rightarrow1$.
Hence, the adaptation becomes slower for larger ratios $\mu/N$.
\begin{figure}[t]
	\begin{subfigure}{0.5\textwidth}
		\centering
		\includegraphics[]{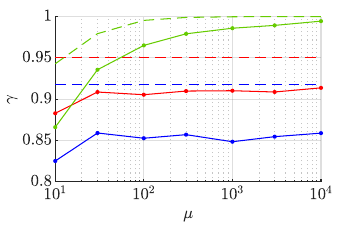}
		\caption{$N=10$.}
		\label{fig:compare_CSAs_a}
	\end{subfigure}
	\begin{subfigure}{0.5\textwidth}
		\centering
		\includegraphics[]{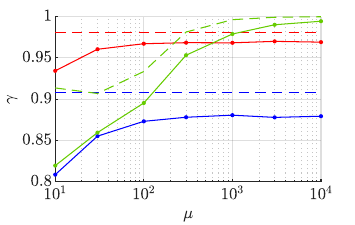}
		\caption{$N=100$}
		\label{fig:compare_CSAs_b}
	\end{subfigure}
	\begin{subfigure}{0.5\textwidth}
		\centering
		\includegraphics[]{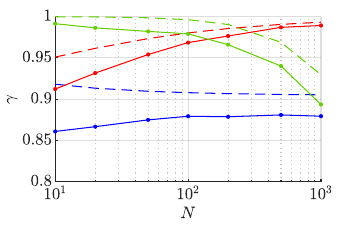}
		\caption{$\mu=1000$.}
		\label{fig:compare_CSAs_c}
	\end{subfigure}
	\begin{subfigure}{0.5\textwidth}
		\centering
		\includegraphics[]{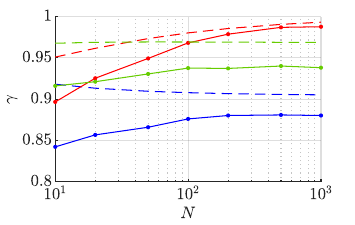}
		\caption{$\mu=2N$.}
		\label{fig:compare_CSAs_d}
	\end{subfigure}
	\caption{Steady-state $\gamma$ for $\muilam{\mu}{\lambda}$-CSA-ES ($\vt=1/2$). 
		Measured ratio $\signss/\sigma^*_0$ (solid, with dots) compared to $\gamma$ from \cref{eq:csa_gamma_b_v2} (dashed) for the CSA variants \cref{eq:sqrtN} (blue), \cref{eq:linN} (red), and \cref{eq:han} (green).
		$\signss$ is determined as follows. First, using $i=1,\dots,M$ trials, the median dynamics $\sigma_M^{*,(g)}$ = $\mathrm{median}(\sigma_i^{*,(g)})$ is determined over all $M$ trials.
		Then, $\signss$ = $\mathrm{median}(\sigma_M^{*,(g_\mathrm{end}/2:g_\mathrm{end})})$ is evaluated over the last generations $g=g_\mathrm{end}/2,\dots,g_\mathrm{end}]$ to reduce  initialization effects.
		$M=5$ at least (e.g.~for $\mu=2000$ and $N=1000$) and $M=100$ the most (e.g. $\mu=N=10$).
		}
	\label{fig:compare_CSAs_gamma}
\end{figure}\noindent
In Fig.~\cref{fig:compare_CSAs_gamma}, experiments on the sphere are conducted to compare the CSA variants with the respective theoretical prediction.
To this end, the dimensionality $N$ and population size $\mu$ are varied.
The measured steady-state $\signss$ is averaged over at least 10 trials
and the median is taken over the measured $\sigma^{*,(g)}$ (the median is necessary due to a slightly skewed distribution of $\sigma^*$ at small $N$).
The measured $\signss$ is normalized by $\sigma^*_0$, yielding a reference value for $\gamma$. 

Figure~\cref{fig:compare_CSAs_gamma} displays the adaptation characteristics of the CSA-variants.
In Figs.~\cref{fig:compare_CSAs_a} and 	\cref{fig:compare_CSAs_b}, the measured ratio remains relatively constant for sufficiently large $\mu$ for \cref{eq:sqrtN} and \cref{eq:linN}.
CSA \cref{eq:han} shows a significant increase of the ratio as $\mu$ increases, which was expected from its damping $D$.
Deviations between the theoretical prediction (dashed) and measurement (solid) can be observed which are mostly due to the large population approximations (see also Sec.~\cref{sec:csa_iteration}).
In Fig.~\cref{fig:compare_CSAs_c}, the dimensionality is varied.
CSA \cref{eq:sqrtN} remains relatively constant, while $\gamma$ of \cref{eq:linN} increases for larger $N$, which was expected from its high damping $D\propto N$.
CSA~\cref{eq:han} shows a decreasing $\gamma$ due to $D\propto \sqrt{\mu/N}$.
In Fig.~\cref{fig:compare_CSAs_d}, both $\mu$ and $N$ are varied together by maintaining $\mu=2N$.
CSA \cref{eq:sqrtN} and \cref{eq:han} remain approximately constant, while for \cref{eq:linN} $\gamma\rightarrow1$.
In general, only \cref{eq:sqrtN} maintains an approximately constant ratio (best agreement for large $N$ and $\mu\gg N$) at $\gamma\lessapprox0.9$, which agrees satisfactory with the prediction \cref{eq:csa_ss_35}.
Furthermore, its adaptation speed is the highest, yielding the lowest $\sigma^*$- and largest $\varphi^*$-levels (cf.~Fig.~\cref{fig:dyn_phi_csa_sa}).
Note that all three CSA yield relatively large ratios between 0.8 and 1.
This means they achieve comparably low progress rates according to Fig.~\cref{fig:phi_sign}.
Furthermore, their $\sigma^*$-levels are not optimal, i.e., they are not maximizing $\varphi^*$ on the sphere.
However, regarding highly multimodal functions with adequate global structure, slow-adaptation is usually beneficial to achieve high success rates.

\section{Mutative Self-Adaptation}
\label{sec:gamma_sa}

\begin{figure}[t]
	\begin{subfigure}{0.5\textwidth}
		\centering
		\includegraphics[]{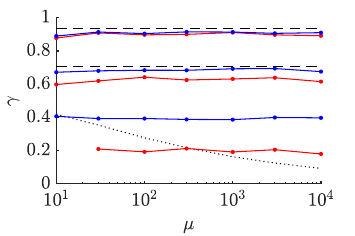}
		\caption{$N=10$.}
		\label{fig:compare_SA_a}
	\end{subfigure}
	\begin{subfigure}{0.5\textwidth}
		\centering
		\includegraphics[]{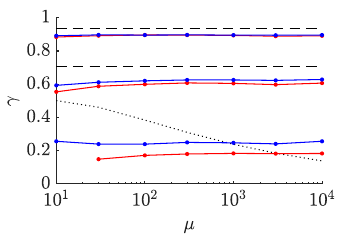}
		\caption{$N=100$}
		\label{fig:compare_SA_b}
	\end{subfigure}
	\begin{subfigure}{0.5\textwidth}
		\centering
		\includegraphics[]{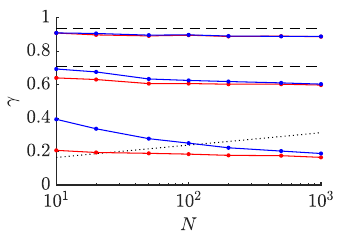}
		\caption{$\mu=1000$.}
		\label{fig:compare_SA_c}
	\end{subfigure}
	\begin{subfigure}{0.5\textwidth}
		\centering
		\includegraphics[]{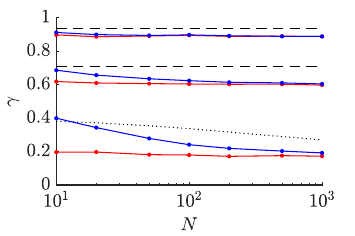}
		\caption{$\mu=2N$.}
		\label{fig:compare_SA_d}
	\end{subfigure}
	\caption{Steady-state $\gamma$ for $\muilam{\mu}{\lambda}$-$\sigma$SA-ES ($\vt=1/2$). 
		Measured ratio $\gamma=\signss/\sigma^*_0$ (solid, with dots) for $\SALN$ (blue), $\SAEP$ (red), showing $\tau=1/\sqrt{8N},1/\sqrt{2N},1/\sqrt{N}$ (top to bottom).
		The dotted black line marks $\hat{\sigma}^*/\sigma^*_0$, and the dashed black lines \cref{eq:sa_phi_psi_eq_exampleGamma1} (bottom) and \cref{eq:sa_phi_psi_eq_exampleGamma2} (top).
		Missing data points of $\SAEP$ at $\mu=10$ are due to adaptation instabilities (see Fig.~\cref{fig:SAEP_instable}).
		The evaluation of $\signss$ es explained in Fig.~\cref{fig:compare_CSAs_gamma}.
	}
	\label{fig:compare_SA_gamma}
\end{figure}

The derivation of the adaptation ratio $\gamma$ for large populations is now continued for $\sigma$SA, see Alg.~\cref{alg:sa}.
The derivation is based on the self-adaptation response function using log-normal mutation sampling.
However, it can also be applied for normal mutation sampling (see below).
For the derivation, the first step is to derive a steady-state condition between the normalized progress rate $\varphi^*$, see \cref{sec:dyn_phidef} and \cref{eq:sign}, and the self-adaptation response function $\psi \coloneqq \EV{(\sigma^{(g+1)}-\sigma^{(g)})/\sigma^{(g)}}$, see \cite[(7.31)]{Bey00b}.
In the sphere steady-state, it holds \cite[(7.162)]{Bey00b}
\begin{align}\label{eq:dyn_R_exp4b}
	\varphi^*(\signss) = -N\psi(\signss,\tau).
\end{align}
Positive progress rates require $\psi<0$, i.e., the expected change of the relative mutation strength is negative as the optimizer is approached.
Equation~\cref{eq:dyn_R_exp4b} was evaluated in \cite[(35)]{OB24} using the large population assumption $\mu\gg N$ and yields
\begin{align}\begin{split}\label{eq:sa_phi_psi_eq}
		\CTHETA\sqrt{2N} - \frac{\sigma_{\mathrm{ss}}^{*2}}{2\mu} = -N\tau^2\qty(\frac12 -\CTHETA\sqrt{2N} +2 e_\vartheta^{1,1}).
\end{split}\end{align}
The condition holds for log-normal sampling of mutation strengths (normal mutation sampling is discussed below).
Given the condition \cref{eq:sa_phi_psi_eq}, one has to insert a suitable steady-state $\signss$.
Analogous to the CSA, the steady-state is characterized w.r.t.~the second zero on the sphere in the limit of large population sizes \cref{eq:signzero_approx_gam_v2}in terms of $\gamma$.
By inserting \cref{eq:signzero_approx_gam_v2} into \cref{eq:sa_phi_psi_eq}, one gets
\begin{align}\begin{split}\label{eq:sa_phi_psi_eq_exp1}
		\CTHETA\sqrt{2N} - \gamma^2\CTHETA\sqrt{2N} &= -N\tau^2\qty(\frac12 -\CTHETA\sqrt{2N} +2 e_\vartheta^{1,1}) \\
		1-\gamma^2 &= N\tau^2\qty(1-\frac{1}{2\CTHETA\sqrt{2N}} -\frac{2 e_\vartheta^{1,1}}{\CTHETA\sqrt{2N}}).
\end{split}\end{align}
For sufficiently large $N$, one can drop the terms of $O(1/\sqrt{2N})$ on the RHS of \cref{eq:sa_phi_psi_eq_exp1}.
Solving for $\tau$ and $\gamma$, respectively, one gets
\begin{align}
	\tau &\simeq \sqrt{\frac{1-\gamma^2}{N}} \label{eq:sa_phi_psi_eq_tau} \\
	\gamma &\simeq \sqrt{1-N\tau^2}. \label{eq:sa_phi_psi_eq_gamma} 
\end{align}
Note that similar results can be obtained if $\SAEP$ is used with normal mutation sampling
\footnote{
	The self-adaptation response function $\psi$ for log-normal and normal mutation sampling differs only by an additional constant bias term $1/2$.
	Details of the derivation can be found in \cite{OB24a}.
	The bias term emerges for log-normal mutations due to the expected value being larger than the initial value, see \cref{eq:metaep_logn2}.
	Hence, the steady-state condition in \cref{eq:sa_phi_psi_eq_exp1} includes a factor $1/2$ on the right-hand side.
	For normal mutation sampling, one obtains analogously 
	\begin{align}\begin{split}\label{eq:saep_phi_psi_eq}
			\CTHETA\sqrt{2N} - \frac{\sigma_{\mathrm{ss}}^{*2}}{2\mu} = -N\tau^2\qty(-\CTHETA\sqrt{2N} +2 e_\vartheta^{1,1}).
	\end{split}\end{align}
	By inserting \cref{eq:signzero_approx_gam_v2} into \cref{eq:saep_phi_psi_eq}, one gets
	\begin{align}\begin{split}\label{eq:sa_phi_psi_eq_exp1b}
			\CTHETA\sqrt{2N} - \gamma^2\CTHETA\sqrt{2N} &= 
			N\tau^2\qty(\CTHETA\sqrt{2N} -2 e_\vartheta^{1,1}) \\
			1-\gamma^2 &= N\tau^2\qty(1-\frac{2 e_\vartheta^{1,1}}{\CTHETA\sqrt{2N}}).
	\end{split}\end{align}
	By neglecting $O(1/\sqrt{2N})$ and solving for $\gamma$, one obtains the same results as in \cref{eq:sa_phi_psi_eq_tau} and \cref{eq:sa_phi_psi_eq_gamma}.
	}.
A few interesting things can be noted from the obtained results.
Given a constant $\gamma$, the learning parameter scales with $1/\sqrt{N}$ in the limit of large population size.
The obtained result agrees with the scaling of the default choice $\tau=1/\sqrt{2N}$ from \cite{Mey07}, which was derived under different assumptions ($N\rightarrow\infty$, $\mu\ll N$).
It also agrees with $\tau=1/\sqrt{N}$ for the $(1,\lambda)$-ES derived in \cite[Sec.~7.4]{Bey00b}.
Furthermore, it agrees with the scaling of the cumulation parameter $\cs$ derived in \cref{eq:csa_ss_32_main}.
Therefore, this scaling appears as a characteristic quantity of $\sigma$-adaptation on the sphere function.
In principle, one can infer $\gamma\in[0,1)$ from \cref{eq:sa_phi_psi_eq_tau}.
As an example, one may evaluate the following $\gamma$-values from \cref{eq:sa_phi_psi_eq_gamma}
\begin{align}
	\gamma &=\sqrt{1-1/2}\approx0.71\qq{for $\tau=1/\sqrt{2N}$} \label{eq:sa_phi_psi_eq_exampleGamma1} \\
	\gamma &=\sqrt{1-1/8}\approx0.94\qq{for $\tau=1/\sqrt{8N}$.} \label{eq:sa_phi_psi_eq_exampleGamma2} 
\end{align}
One may also set $\tau=1/\sqrt{N}$ ($\gamma=0$) in experiments and it yields fast adaptation rates (see Fig.~\cref{fig:compare_SA_gamma}).
However, it is clear that the approximation quality of the progress rate deteriorates as $\gamma\rightarrow 0$ and for increasing $\tau$, see discussion of Fig.~\cref{fig:dyn_phi_csa_sa}.
In this case, the theoretical prediction of $\gamma$ does not yield useful results.
However, in experiments, the actually realized $0.2 \lessapprox\gamma_\mathrm{meas}\lessapprox0.4$ yields fast (but potentially unstable) $\sigma$-adaptation.
\noindent
Figure~\cref{fig:compare_SA_gamma} investigates the adaptation properties of the $\sigma$SA-ES using log-normal mutations and normal mutations, respectively, see Alg.~\cref{alg:sa}.
The same configurations as in Fig.~\cref{fig:compare_CSAs_gamma} are tested, however, note that the $y$-axis scale is different.
Furthermore, the optimal ratio $\hat{\sigma}^*/\sigma^*_0$ (maximizing \cref{eq:sph_Ndep_pc}) is now additionally displayed since it appears at comparably small $\gamma$-values.
The three tested $\tau$-values can be categorized as slow ($1/\sqrt{8N}$), default ($1/\sqrt{2N}$), and fast adaptation ($1/\sqrt{N}$) on the sphere function.
The differences between $\SALN$ and $\SAEP$ are negligible at small $\tau$ and increase with larger $\tau$.
It can be observed that $\SAEP$ realizes somewhat smaller values of $\gamma$ (depending on $\tau$ and $N$).
This can be attributed to the fact that the sampling is unbiased.
Due to the bias, $\SALN$ realizes slightly larger mutation strength levels and the differences vanish for $\tau\rightarrow\infty$.
For constant $\mu$ and varying $N$ (top plots), both $\sigma$SA realize an approximately constant $\gamma$-level.
With increasing $N$ (bottom plots), a slight downward trend of $\gamma$ can be observed.
For smaller $\tau$-values (slower adaptation), this effect is relatively small.
Interestingly, the $\sigma$SA realizes a large range of possible $\gamma$-levels depending on the chosen $\tau$, reaching from (approximately) 0.2 to 0.95.
In contrast to that, the CSA $\gamma$-levels from Fig.~\cref{fig:compare_CSAs_gamma} lie above 0.8 for all standard CSA-implementations.

\begin{figure}[t]
	\centering
	\includegraphics[width=\mywidth]{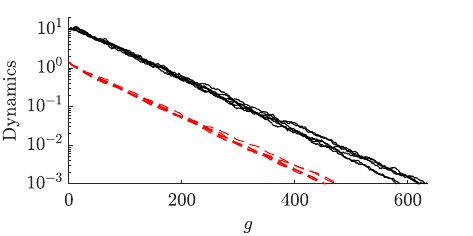}
	\includegraphics[width=\mywidth]{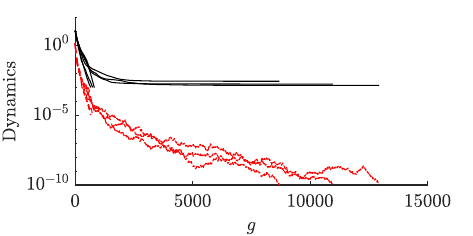}
	\caption{Dynamics ($R^{(g)}$: solid black, $\sigma^{(g)}$: dashed red) of six trials of $\muilam{10}{20}$-$\sigma$SA-ES on the sphere for $N=100$ at $\tau=1/\sqrt{N}$ with $\SALN$ (left) and $\SAEP$ (right: three trials showing $\sigma<\sigma_\mathrm{stop}$).
	}
	\label{fig:SAEP_instable}
\end{figure}

One important observation was made for $\SAEP$ in Fig.~\cref{fig:compare_SA_gamma}.
Note that for $\tau=1/\sqrt{N}$, one data point is missing at $\mu=10$.
The $\SAEP$ with normal mutations becomes unstable for large $\tau$ and small $\mu$.
Example dynamics are shown in Fig.~\cref{fig:SAEP_instable}.
On the left, $\SALN$ achieves convergence of all trials by reaching the target
$R_\mathrm{stop}<10^{-3}$.
On the other hand, some runs of the $\SAEP$ become unstable and reach the $\sigma$-stopping criterion $\sigma_\mathrm{stop}<10^{-10}$ before reaching $R_\mathrm{stop}$.
As an example, for $\mu=10$ at $N=100$ in Fig.~\cref{fig:compare_SA_b}, 7 out of 100 runs reach $\sigma_\mathrm{stop}$.
Since this is an undesired (unstable) behavior, $\gamma$ is not evaluated.
On the other hand, the bias of $\SALN$ helps to keep $\sigma$ stable at larger $\tau$-values.
However, this example also illustrates that $\tau$ cannot be arbitrarily increased.

%% file: conclusion.tex
\section{Conclusion}
\label{sec:conc}

In this paper, standard implementations of the CSA-ES and self-adaptive $\sigma$ES were investigated for a multi-recombinative $\muilam{\mu}{\lambda}$-ES with isotropic mutations of strength $\sigma$.
The goal was to investigate the adaptation on the sphere as a function of the population size ($\mu$) and search space dimensionality ($N$).
To this end, the steady-state scale-invariant mutation strength $\sigma^*$ was studied.
An approach was presented to characterize the adaptation speed in terms of its steady-state $\signss$-level in relation to the maximum $\sigma^*$-value that yields positive progress on the sphere.
To this end, an adaptation scaling factor $\gamma\in(0,1)$ was introduced with $\gamma\rightarrow1$ denoting slowest possible adaptation due to vanishing progress.
Then, analytic investigations were carried out to predict $\gamma$ as a function of $\mu$ and $N$.
To this end, steady-state solutions of the CSA and $\sigma$SA on the sphere function were derived under certain simplifying assumptions.
The results showed satisfactory agreement with experiments, which was expected from the applied approximations.

The analysis of the CSA has revealed largely different adaptation properties of standard CSA-implementations as functions of $\mu$ and $N$.
Only the CSA~\cref{eq:sqrtN} (cumulation $\cs=1/\sqrt{N}$ and damping $D=\cs^{-1}$) shows (approximately) constant scaling of $\gamma(\mu,N)$.
In this case, changing $\mu$ or $N$ does not significantly impact its adaptation in terms of $\gamma$, which is desirable.
Other CSA-variants show significant changes in $\gamma$ when $\mu$ or $N$ are varied.
As an example, CSA~\cref{eq:linN} becomes increasingly slower for large $N$, while \cref{eq:han} becomes slower for larger ratios $\mu/N$.
Analyzing $\sigma$SA has illustrated the progress rate decrease for increasing learning rate $\tau$.
However, depending on $\tau$, its adaptation has shown to reach largely different $\signss$-levels compared to the CSA, which can lead to significantly higher progress rates on the sphere.
Furthermore, it also maintains relatively constant $\gamma$-levels as a function of $\mu$ and $N$.
While differences exist between $\SALN$ and $\SAEP$ (especially for small $N$ and $\mu$), they are negligible for sufficiently small $\tau$.

The obtained results will be useful for future investigations of adaptive population control for ES. 
It is clear that a better understanding of the $\sigma$-adaptation as a function of the population size will also help to understand the search behavior of ES when the population size changes dynamically.